\documentclass[10pt,twocolumn,letterpaper]{article}

\usepackage{iccv}
\usepackage{times}
\usepackage{epsfig}
\usepackage{graphicx}
\usepackage{amsmath}
\usepackage{amssymb}
\usepackage{cite}
\usepackage{mathrsfs}
\usepackage{color}
\usepackage[accsupp]{axessibility}  
\usepackage{url}

\usepackage[breaklinks=true,bookmarks=false]{hyperref}

\iccvfinalcopy 

\def\iccvPaperID{****} 

\ificcvfinal\fi

\begin{document}

\title{ISNet: Integrate Image-Level and Semantic-Level Context \\for Semantic Segmentation}

\author{Zhenchao Jin,~ Bin Liu\thanks{Corresponding author.},~ Qi Chu,~ Nenghai Yu\\
CAS Key Laboratory of Electromagnetic Space Information,\\ University of Science and Technology of China \\
{\tt\small\{blwx@mail., flowice@, qchu@, ynh@\}ustc.edu.cn }
}

\maketitle
\ificcvfinal\thispagestyle{empty}\fi

\begin{abstract}
   Co-occurrent visual pattern makes aggregating contextual information a common paradigm to enhance the pixel representation for semantic image segmentation.
   The existing approaches focus on modeling the context from the perspective of the whole image, i.e., aggregating the image-level contextual information.
   Despite impressive, these methods weaken the significance of the pixel representations of the same category, i.e., the semantic-level contextual information.
   To address this, this paper proposes to augment the pixel representations by aggregating the image-level and semantic-level contextual information, respectively.
   First, an image-level context module is designed to capture the contextual information for each pixel in the whole image.
   Second, we aggregate the representations of the same category for each pixel where the category regions are learned under the supervision of the ground-truth segmentation.
   Third, we compute the similarities between each pixel representation and the image-level contextual information, the semantic-level contextual information, respectively.
   At last, a pixel representation is augmented by weighted aggregating both the image-level contextual information and the semantic-level contextual information with the similarities as the weights.
   Integrating the image-level and semantic-level context allows this paper to report state-of-the-art accuracy on four benchmarks, i.e., ADE20K, LIP, COCOStuff and Cityscapes \footnote{Our code will be available at \href{https://github.com/SegmentationBLWX/sssegmentation}{https://github.com/SegmentationBLWX\\/sssegmentation}.}.
\end{abstract}

\vspace{-0.50cm}

\section{Introduction}

Semantic image segmentation, which assigns per-pixel predictions of object categories for the given image, is a fundamental problem in computer vision.
This task is exceptionally significant to tons of real-world applications, \emph{e.g.}, automatic driving and robot sensing.
Recent developments of deep neural networks \cite{simonyan2014very,he2016deep} encourage the emergence of a series of works \cite{chen2017deeplab,chen2017rethinking,zhao2017pyramid,chen2018encoder,long2015fully,yuan2019object}, where FCN is the cornerstone of these works.
Its encoder-decoder structure, which reduces the spatial dimension to extract features and then leverages upsampling to recover spatial extent, shows numerous improvements in semantic segmentation.
Based on this, recent researches mainly focus on two issues to further boost the segmentation performance.
One is how to improve the encoder structure so that the models can extract more robust representation for each pixel \cite{yu2017dilated,wang2020deep,zhang2020resnest}.
The other is how to model the context so that the network can enhance the representation capability of each pixel by encoding the contextual information into the original feature representations \cite{fu2019dual,zhao2017pyramid,chen2017deeplab,yuan2019object,he2019dynamic,huang2019ccnet}, which is also the interest of this paper.

\begin{figure}
\centering
\includegraphics[width=0.45\textwidth]{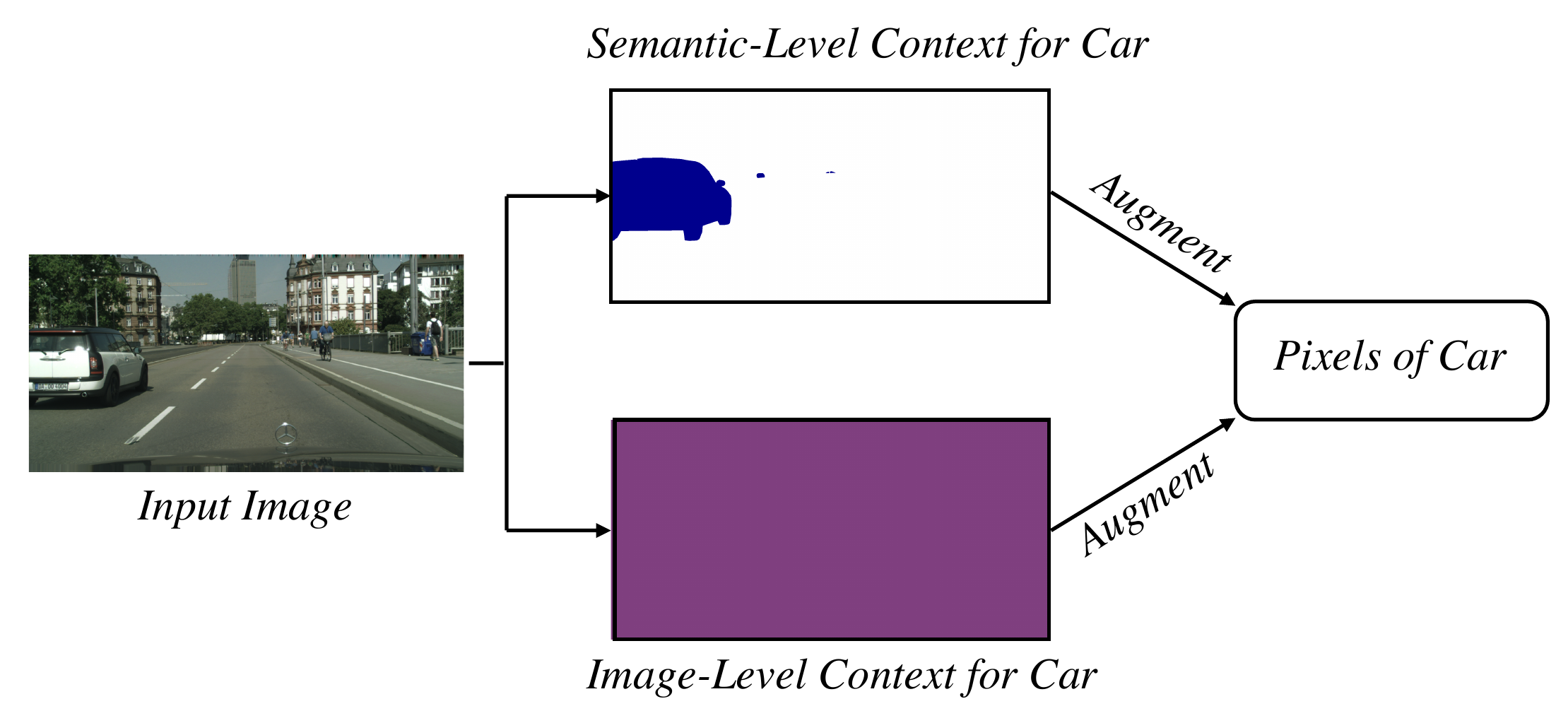}
\caption{
   Main idea of integrating image-level and semantic-level context.
   The \textcolor{blue}{blue} region on the upper branch denotes for semantic-level context and the \textcolor[rgb]{0.5,0,0.5}{purple} area on the lower branch stands for image-level context.
}\label{motivation}
\vspace{-0.50cm}
\end{figure}

Co-occurrent visual pattern inspires the emergence of a series of works about modeling context.
These approaches can be roughly divided into two types, \emph{i.e.}, multi-scale context modeling and relational context modeling.
For multi-scale context modeling, Deeplab \cite{chen2017deeplab} introduces the atrous spatial pyramid pooling (ASPP) so that it can leverage various dilation convolutions to capture the multi-scale contextual information.
PSPNet \cite{zhao2017pyramid} proposes to utilize pyramid spatial pooling to aggregate the multi-scale contextual information.
For relational context modeling, Wang \emph{et al.} \cite{wang2018non} first revisit the traditional local means \cite{buades2005non}, and then design a non-local block to weighted aggregate the contextual information in the whole image.
Zhu \emph{et al.} propose an asymmetric pyramid non-local block to decrease the compution and GPU memory consumption of the standard non-local module.
Apart from this, ACFNet \cite{zhang2019acfnet} and OCRNet \cite{yuan2019object} first group the pixels into a set of regions, and then augment the pixel representations by weighted aggregating the region representations where the weights are determined by the relations between the pixels and regions.
Though impressive, these solutions only focus on aggregating the contextual information from the perspective of the whole image (\emph{i.e.}, image-level contextual information), while discard the significance of the pixel representations of the same category.
Accordingly, they all suffer from the same issue that the contextual information of each pixel is unevenly captured from the category region the pixel belongs to and the regions of other categories.
For instance, the pixels in the boundaries or the regions of objects of small scales tend to capture much more contextual information from the regions of other objects.
Since the label of a pixel is the category of the object that the pixel belongs to, too much contextual information from other objects may cause the network to mislabel these pixels as other categories.

To alleviate the problem above, this paper proposes to augment the pixel representations by aggregating the image-level and semantic-level contextual information, respectively.
As illustrated in Figure \ref{motivation}, the image-level context stands for all the pixels in the input image and the semantic-level context denotes for the pixels in the same category region. 
Based on this definition, an image-level context module (ILCM) is first designed to capture the contextual information from the whole image and thereby, we can obtain the image-level contextual information.
Then, a novel semantic-level context module (SLCM) is proposed to aggregate representations of the same category for each pixel (\emph{i.e.}, the semantic-level contextual information), where the category regions are learned under the supervision of the ground-truth segmentation.
Next, the similarities between the pixel representation and the image-level contextual information, the semantic-level contextual information are calculated.
At last, the pixel representations are augmented by weighted aggregating the image-level contextual information and the semantic-level contextual information, 
where the weights are determined by the calculated similarities.
On the whole, our major contributions are summarized as follows:

\begin{itemize}

\item To the best of our knowledge, this paper first explores improving the pixel representations by aggregating the image-level contextual information and semantic-level contextual information, respectively.

\item This paper designs a simple yet effective image-level context module (ILCM) and a novel semantic-level context module (SLCM) to capture the contextual information from the perspective of the whole image and the category region, respectively.
Experimental results demonstrate the effectiveness of our method.

\item A general architecture framework named ISNet is proposed in this paper, which reveals how to leverage ILCM and SLCM to consistently boost the performance of semantic image segmentation. 
The proposed framework allows this paper to achieve state-of-the-art accuracy on four segmentation benchmarks, \emph{i.e.}, ADE20K, LIP, COCOStuff and Cityscapes.

\end{itemize}

\section{Related Work}

\noindent \textbf{Semantic Segmentation.}
To generate pixel-wise semantic predictions for a given image, image classification networks \cite{chen2017deeplab,simonyan2014very} are extended to yield semantic segmentation masks.
FCN \cite{long2015fully} is the first work to apply fully convolution on the whole image to produce labels of every pixel and many researchers have made efforts based on FCN in the past few years.
Specifically, these studies can be roughly divided into two groups. 
One is to design a novel backbone network \cite{wang2020deep,yu2017dilated} to extract more robust feature representation for each pixel.
Considering high-resolution representations are essential for position-sensitive vision problems, Wang \emph{et al.} \cite{wang2020deep} propose a backbone network named HRNet to maintain high-resolution representations through the whole process.
ResNeSt \cite{zhang2020resnest} presents a modularized architecture to capture cross-feature interactions and learn diverse representations by utilizing the channel-wise attention on different network branches.
The other is to introduce richer contextual information for each pixel \cite{he2019dynamic,ruan2019devil,zhao2017pyramid,chen2017deeplab,cao2019gcnet,fu2019dual}.
For instance, adopting different sizes of convolutional/pooling kernels or dilation rates to gather multi-scale visual cues \cite{zhao2017pyramid,chen2017deeplab,yang2018denseaspp}, 
employing neural attention \cite{chen2016attention} to directly exchange the contextual information between paired pixels \cite{huang2019ccnet,cao2019gcnet,fu2019dual,yin2020disentangled,yuan2018ocnet}
and building the image pyramids or feature pyramids \cite{kirillov2019panoptic,liu2015parsenet}.
This paper focuses on the latter one, \emph{i.e.}, aggregating more meaningful contextual information to augment the pixel representations.

\begin{figure*}
\centering
\includegraphics[width=0.85\textwidth]{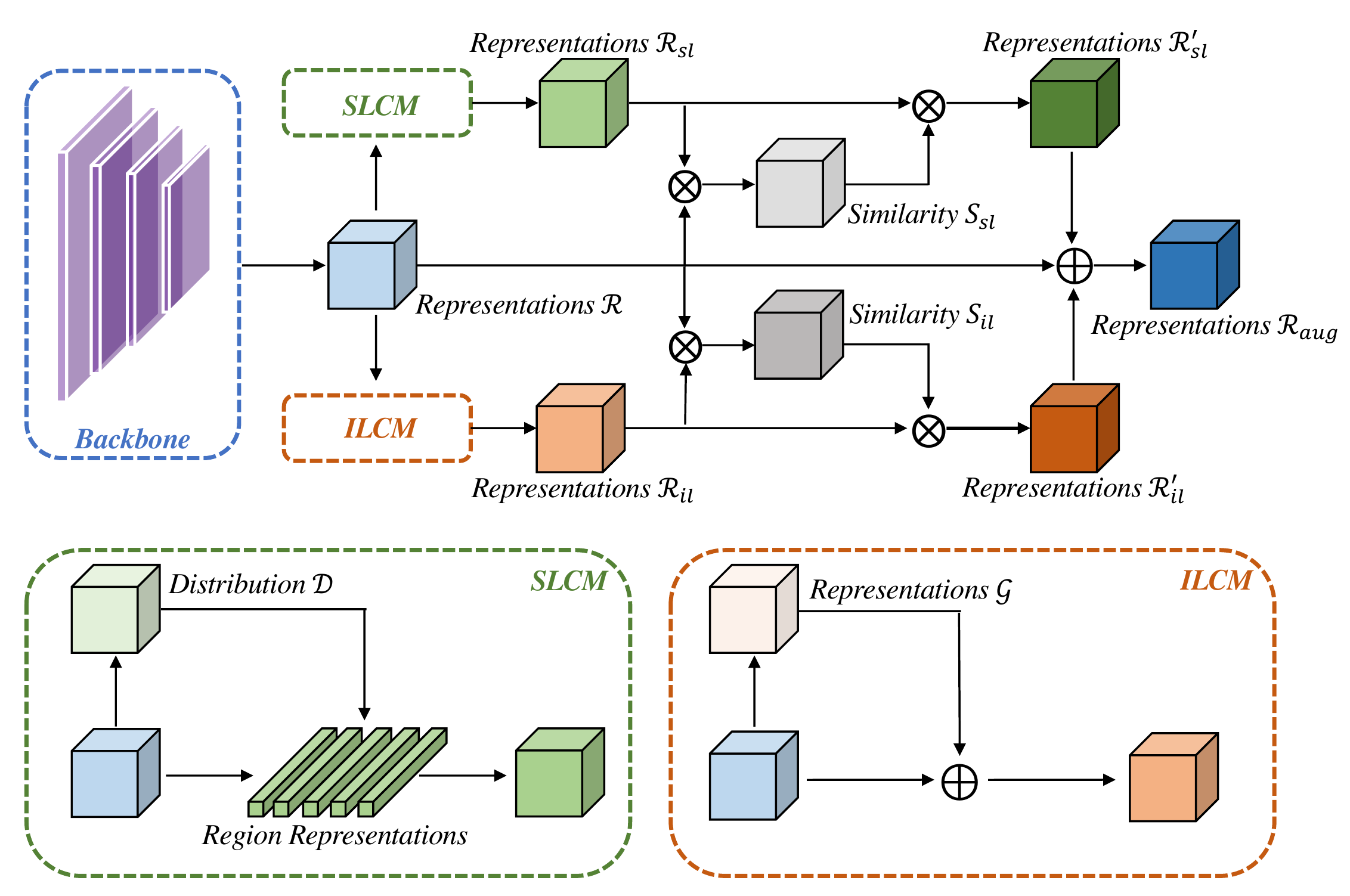}
\caption{
   The overview of the proposed framework (ISNet).
   First, the semantic-level context module (SLCM) and image-level context module (ILCM) are utilized to extract the semantic-level contextual information $\mathcal{R}_{sl}$ and the image-level contextual information $\mathcal{R}_{il}$, respectively.
   Then, we calculate the similarities between pixel representations $\mathcal{R}$ and $\mathcal{R}_{sl}$, $\mathcal{R}_{il}$.
   At last, both contextual information are adopted to augment the pixel representations according to the calculated similarities.
}\label{framework}
\vspace{-0.30cm}
\end{figure*}

\noindent \textbf{Context Aggregation.}
While FCN captures information from bottom-up, contextual information with wide field-of-view is also critical for pixel labeling task and is exploited by numerous studies.
Deeplab \cite{chen2017deeplab, chen2017rethinking, chen2018encoder} proposes atrous convolution kernels to force the network to perceive larger area and obtain the higher resolution output. 
Based on Deeplab, DenseASPP \cite{yang2018denseaspp} densifies the dilated rates to make the network conver larger scale ranges.
PSPNet \cite{zhao2017pyramid} adopts spatial pooling to gain the feature maps of different receptive field sizes so that the combined feature maps could aggregate multi-scale object clues.
DANet \cite{fu2019dual} and OCNet \cite{yuan2018ocnet} first calculate the similarities between the pixels as the weights, and then improve the pixel representations by weighted aggregation of all the pixels in the input image.
Apart from this, some works \cite{yuan2019object,zhang2019acfnet,li2019expectation} first group the pixels into a set of regions, and then the pixel representations are augmented by weighted aggregating the region representations where the weights are determined by their context relations.
Although our semantic-level context module sees similar to these methods, the key difference is that we augment a pixel representation only by adopting the region representation with the same category as the pixel representation rather than all region representations.

\section{Methodology}
As demonstrated in Figure \ref{framework}, our ISNet incorporates the image-level contextual information and semantic-level contextual information for semantic image segmentation.
We first introduce the overall formulation of our framework in Section \ref{sec3.1}.
Then, the details of image-level context module (ILCM) and semantic-level context module (SLCM) are described in Section \ref{sec3.2} and Section \ref{sec3.3}, respectively.
Finally, we show the multi-task loss function used for training the proposed ISNet in Section \ref{sec3.4}.

\subsection{Formulation}\label{sec3.1}
Given the input image $\mathcal{I} \in \mathbb{R}^{3 \times H \times W}$, 
we first use a backbone network $\mathscr{B}$ (\emph{e.g.}, ResNet \cite{he2016deep}) to project the pixels in $\mathcal{I}$ into a non-linear embedding space so that we can obtain the pixel representations $\mathcal{R}$:
\begin{equation}
   \mathcal{R} = \mathscr{B}(\mathcal{I}),
\end{equation}
where $\mathcal{R}$ is a matrix of size $C \times \frac{H}{8} \times \frac{W}{8}$ and the dimension of a pixel representation is $C$.

Then, the image-level context module $\mathscr{M}_{il}$ is utilized to aggregate the contextual information from the whole image:
\begin{equation}\label{eq3}
   \mathcal{R}_{il} = \mathscr{M}_{il}(\mathcal{R}),
\end{equation}
where $\mathcal{R}_{il}$ is a matrix of size $C \times \frac{H}{8} \times \frac{W}{8}$ storing the image-level contextual information for each pixel representation.
Simultaneously, the semantic-level context module $\mathscr{M}_{sl}$ is designed to capture the contextual information within individual category regions:
\begin{equation}\label{eq4}
   \mathcal{R}_{sl} = \mathscr{M}_{sl}(\mathcal{R}),
\end{equation}
where $\mathcal{R}_{sl}$ is a matrix of size $C \times \frac{H}{8} \times \frac{W}{8}$ storing the semantic-level contextual information for each pixel representation.
After that, we calculate the similarities between $\mathcal{R}$ and $\mathcal{R}_{il}$:
\begin{equation}\label{eq5}
   \mathcal{S}_{il} = Softmax(\frac{\mathcal{R}^{\frac{HW}{64} \times C} \otimes \mathcal{R}^{C \times \frac{HW}{64}}_{il}}{\sqrt{C}}),
\end{equation}
where $\mathcal{S}_{il}$ is a matrix of size $\frac{HW}{64} \times \frac{HW}{64}$ and $\otimes$ stands for matrix multiplication.
The same for $\mathcal{R}$ and $\mathcal{R}_{sl}$:
\begin{equation}\label{eq6}
   \mathcal{S}_{sl} = Softmax(\frac{\mathcal{R}^{\frac{HW}{64} \times C} \otimes \mathcal{R}^{C \times \frac{HW}{64}}_{sl}}{\sqrt{C}}),
\end{equation}
where $\mathcal{S}_{sl}$ is a matrix of size $\frac{HW}{64} \times \frac{HW}{64}$.
This operation is inspired by the self-attention mechanism \cite{chen2016attention}.
Next, we leverage $\mathcal{R}_{il}$ and $\mathcal{R}_{sl}$ to augment $\mathcal{R}$:
\begin{equation}\label{eq7}
   \mathcal{R}_{aug} = \mathscr{A} (\mathcal{R}'_{il} \oplus \mathcal{R}'_{sl} \oplus \mathcal{R}),
\end{equation}
where $\oplus$ denotes for the concatenation operation and $\mathscr{A}$ is a transform function used to reduce the channels of the input matrix tensors to have size $C \times \frac{H}{8} \times \frac{W}{8}$.
$\mathcal{R}'_{il}$ is calculated by using $\mathcal{S}_{il}$ and $\mathcal{R}_{il}$:
\begin{equation}\label{eq8}
   \mathcal{R}'_{il} = reshape(\mathcal{S}^{\frac{HW}{64} \times \frac{HW}{64}}_{il} \otimes \mathcal{R}^{\frac{HW}{64} \times C}_{il}),
\end{equation}
and $\mathcal{R}'_{sl}$ is obtained as follows:
\begin{equation}\label{eq9}
   \mathcal{R}'_{sl} = reshape(\mathcal{S}^{\frac{HW}{64} \times \frac{HW}{64}}_{sl} \otimes \mathcal{R}^{\frac{HW}{64} \times C}_{sl}),
\end{equation}
where $reshape$ is used to make $\mathcal{R}'_{il}$ and $\mathcal{R}'_{sl}$ have size of $C \times \frac{H}{8} \times \frac{W}{8}$.
At last, $\mathcal{R}_{aug}$ is leveraged to predict the labels of the pixels in $\mathcal{I}$:
\begin{equation}\label{eq10}
   \mathcal{O} = Upsample_{8\times} (\mathscr{H}(\mathcal{R}_{aug})),
\end{equation}
where $\mathscr{H}$ is a classification head and $\mathcal{O}$ is a matrix of size $K \times H \times W$ storing the predicted class probability distribution of each pixel. $K$ is the number of the categories.

\subsection{Image-Level Context Module}\label{sec3.2}
Image-level context module $\mathscr{M}_{il}$ is designed to capture the contextual information from the perspective of the whole image.
Since there exist co-occurrent visual patterns \cite{zhao2017pyramid}, $\mathscr{M}_{il}$ is widely applied in semantic segmentation task.
Prior to this paper, there have been many excellent structures of $\mathscr{M}_{il}$ such as ASPP \cite{chen2017deeplab}, PPM \cite{zhao2017pyramid} and OCR \cite{yuan2019object}.
Despite this, since there are two context modules in our framework, we expect the designed $\mathscr{M}_{il}$ in this paper owning the least computation complexity and the increased parameters.
Following this expectation, as indicated in Figure \ref{framework}, we first calculate the channel-wise mean values of the matrix tensor $\mathcal{R}$:
\begin{equation}\label{eq11}
   \mathcal{G} = \frac{1}{\frac{H}{8} \times \frac{W}{8}} \sum_{ij}\mathcal{R}_{[*, i, j]},
\end{equation}
where we leverage the subscript $[i, j]$ or $[*, i, j]$ to index the element or elements of a matrix.
$\mathcal{G}$ is a matrix of size $C \times 1 \times 1$ storing the global contextual information of corresponding channels.
Then, $\mathcal{G}$ is added into the pixel representations $\mathcal{R}$ to obtain $\mathcal{R}_{il}$:
\begin{equation}\label{eq12}
   \mathcal{R}_{il} = \mathscr{F} (repeat(\mathcal{G}) \oplus \mathcal{R}),
\end{equation}
where $\mathscr{F}$ is a transform function used to fuse $\mathcal{G}$ and $\mathcal{R}$, implemented by a $1 \times 1$ convolutional layer.
$repeat$ is used to repeat the elements in the corresponding channels of $\mathcal{G}$ to make $\mathcal{G}$ have the same shape as $\mathcal{R}$. 
Note that, the image-level context module can be replaced by all of the existing method such as ASPP \cite{chen2017deeplab} and PPM \cite{zhao2017pyramid} for better modeling the image-level context when pursuing the best segmentation performance.
It does not affect the basic motivation of this paper.

\subsection{Semantic-Level Context Module}\label{sec3.3}
Semantic-level context module $\mathscr{M}_{sl}$ is proposed to aggregate the contextual information within individual category regions.
As shown in Figure \ref{framework}, a classification head $\mathscr{H}'$ is first introduced to predict the category probability distribution $\mathcal{D}$ of the representations in $\mathcal{R}$:
\begin{equation}\label{eq13}
   \mathcal{D} = \mathscr{H}'(\mathcal{R}),
\end{equation}
where the size of $\mathcal{D}$ is $K \times \frac{H}{8} \times \frac{W}{8}$ and $\mathscr{H}'$ is implemented by two $1 \times 1$ convolutional layers.
According to $\mathcal{D}$, the representations in $\mathcal{R}$ can be grouped into various category regions:
\begin{equation}\label{eq14}
   \mathcal{R}_{c_k} = \{ \mathcal{R}_{[*,i,j]}  ~|~ argmax(\mathcal{D}_{[*,i,j]}) = c_k  \},
\end{equation}
where $c_k$ is between $1$ and $K$ standing for the category label and $\mathcal{R}_{c_k}$ is a matrix of size $N_{c_k} \times C$. $N_{c_k}$ denotes for the number of representations belonging to category $c_k$.
For the convenience of presentation, we also define the $\mathcal{D}_{c_k}$ as:
\begin{equation}\label{eq15}
   \mathcal{D}_{c_k} = \{ \mathcal{D}_{[c_k,i,j]}  ~|~ argmax(\mathcal{D}_{[*,i,j]}) = c_k  \},
\end{equation}
where $\mathcal{D}_{c_k}$ is a matrix of size $N_{c_k} \times 1$.
Next, to aggregate the semantic-level contextual information for each pixel representation according to their category, 
we calculate the region representation for each semantic class $c_k$ as follows:
\begin{equation}\label{eq16}
   \mathcal{R}'_{c_k} = \sum_{n=1}^{N_{c_k}} \frac{e^{\mathcal{D}_{c_k, [n,*]}}}{\sum e^ {\mathcal{D}_{c_k}}} \cdot \mathcal{R}_{c_k, [n, *]}
\end{equation}
where $\mathcal{R}'_{c_k}$ of size $1 \times C$ is the composite vector of the representations of the same category.
After calculating all region representations, we assign them to a matrix tensor according to the class label of corresponding element:
\begin{equation}\label{eq17}
   \mathcal{R}_{sl, [*,i,j]} = \mathcal{R}'_{c_k} ~if~argmax(\mathcal{D}_{[*,i,j]}) = c_k
\end{equation}
where $\mathcal{R}_{sl}$ of size $C \times \frac{H}{8} \times \frac{W}{8}$ is the semantic-level contextual information we demand.

\subsection{Loss Function}\label{sec3.4}
A multi-task loss function of $\mathcal{D}$ and $\mathcal{O}$ is used to jointly optimize the model parameters.
In particular, the loss function of $\mathcal{D}$ is defined as:
\begin{equation}
   \mathcal{L}_{\mathcal{D}} = \frac{1}{H \times W} \sum_{i, j} L_{ce} (\mathcal{D}^{K \times H \times W}_{[*, i,j]}, ~\S (\mathcal{GT}_{[ij]})),
\end{equation}
where $\S$ denotes for converting the ground truth class label stored in $\mathcal{GT}$ into one-hot format and $\mathcal{D}^{K \times H \times W}$ is calculated as follows:
\begin{equation}
   \mathcal{D}^{K \times H \times W} = Softmax(Upsample_{8 \times}(\mathcal{D})).
\end{equation}
$L_{ce}$ denotes for the \emph{cross entropy loss} and $\sum_{i, j}$ denotes that the summation is calculated over all locations on the input image $I$.
To let $\mathcal{O}$ contain the accurate category probability distribution of each pixel, we define the loss function of $\mathcal{O}$ as follows:
\begin{equation}
   \mathcal{L}_{\mathcal{O}} = \frac{1}{H \times W} \sum_{i, j} L_{ce} (\mathcal{O}_{[*, i, j]}, ~\S (\mathcal{GT}_{[ij]})).
\end{equation}
Finally, we formulate the multi-task loss function $\mathcal{L}$ as:
\begin{equation}\label{eq18}
   \mathcal{L} = \alpha \mathcal{L}_{\mathcal{D}} + \mathcal{L}_{\mathcal{O}}
\end{equation}
where $\alpha$ is the hyper-parameters to balance the loss of $\mathcal{L}_{\mathcal{D}}$ and $\mathcal{L}_{\mathcal{O}}$.
We empirically set $\alpha=0.4$ by default.
With this joint loss function, the model parameters are learned jointly through back propagation.

\section{Experiments}
\subsection{Experimental Setup}
\noindent \textbf{Benchmarks.} We conduct the experiments on four widely-used semantic segmentation benchmarks.

\begin{itemize}
\item \textbf{ADE20K} \cite{zhou2017scene} is a scene parsing dataset including 150 categories and diverse scenes with 1,038 image-level labels.
This challenging benchmark is divided into 20K/2K/3K images for training, validation and testing.

\item \textbf{COCOStuff} \cite{caesar2018coco} is a challenging scene parsing dataset that provides rich annotations for 91 thing classes and 91 stuff classes.
The dataset contains 9K/1K images for training and testing, respectively.

\item \textbf{LIP} \cite{gong2017look} is a large-scale benchmark that focuses on single human parsing. It contains 50,426 single-person images, which are divided
into 30,426 images for training, 10,000 for validation and 10,000 for testing. The pixel-wise annotations cover 19 semantic human part labels and one background label.

\item \textbf{Cityscapes} \cite{cordts2016cityscapes} is a benchmark for semantic urban scene understanding that contains 19 semantic classes.
There are 5K high quality pixel-level finely annotated images and 20K coarsely annotated images 
in the dataset. The finely annotated 5K images are divided into sets with numbers 2,975, 500, 1,525 for training, validation and testing.
\end{itemize}

\noindent \textbf{Training Details.} 
We initialize the backbone network using the weights pre-trained on ImageNet and two integrated context modules are initialized randomly.
``Poly'' learning rate policy with factor $(1 - \frac{iter}{total\_iter})^{0.9}$ is performed for training our framework.
Synchronized batch normalization implemented by pytorch is enabled during training.
And for the data augmentation, we augment each sample with random scaling in the range of $[0.5, 2]$, random cropping and left-right flipping during training.
More specifically, following previous works \cite{yuan2019object}, the training settings for different benchmarks are listed as follows:

\begin{itemize}
 
   \item \textbf{ADE20K:} The initial learning rate is set as $0.01$ and the weight decay is $0.0005$.
   The crop size of the input image is set as $512 \times 512$ and batch size is set as $16$ by default.
   The models are fine-tuned for 160K iterations if not specified.
   
   \item \textbf{COCOStuff:} We set the initial learning rate as $0.001$, weight decay as $0.0001$, crop size as $512 \times 512$, batch size as $16$ and training iterations as 60K by default.
 
   \item \textbf{LIP:} The initial learning rate is set as $0.01$ and the weight decay is $0.0005$. The crop size of the input image is set as $473 \times 473$ and batch size is set as $32$ by default.
   If not specified, the models are fine-tuned for 160K iterations.

   \item \textbf{Cityscapes:} The initial learning rate is set as $0.01$ and the weight decay is $0.0005$.
   We set the crop size of the input image as $512 \times 1024$, batch size as $8$ and the training iterations as 80K if not specified.
 
\end{itemize}

\noindent \textbf{Inference Settings.} 
For ADE20K, COCOStuff and LIP, the size of the input image during testing is the same as the size of the input image during training.
And for Cityscapes, the input image is zoomed to have its shorter side being $1024$ pixels. 
By default, no tricks (\emph{e.g.}, multi-scale with flipping testing) will be adopted during testing.

\noindent \textbf{Evaluation Metrics.} 
Following the standard setting, mean intersection-over-union (mIoU) is adopted for evaluation.

\noindent \textbf{Reproducibility.} The proposed framework is implemented on PyTorch ($version \geq 1.3$) and trained on four NVIDIA Tesla V100 GPUs with a 32 GB memory per-card.
And all the testing procedures are performed on a single NVIDIA Tesla V100 GPU. To provide full details of our framework, our code will be made publicly available.

\subsection{Ablation Study}

\noindent \textbf{ILCM.} 
Since there exist co-occurrent visual patterns, the image-level contextual information is significant for semantic image segmentation.
For instance, the car is likely to be in the parking lot or on the highway while not fly in sky.
From Table \ref{table1}, we can see that the image-level contextual module (ILCM) brings an improvement of $5.54\%$ mIoU on the validation set of ADE20K.
This result demonstrates that ILCM owns the ability of modeling the context from the perspective of the whole image and 
so that it helps the network better classify the pixels by considering the long-range dependencies.

\begin{table}[t]
\centering
\caption{
   Ablation experiments on the image-level context module (ILCM) and the semantic-level context module (SLCM).
   All methods are learned on the train set of ADE20K and evaluated using single scale test protocal on the validation set.
}\label{table1}
\resizebox{.4\textwidth}{!}{
\begin{tabular}{ccc|c|c}
   \hline
   \hline
   Baseline         &ILCM            &SLCM             &Backbone           &mIoU      \\
   \hline
   \checkmark       &                &                 &ResNet-50          &36.96     \\
   \checkmark       &\checkmark      &                 &ResNet-50          &42.50     \\
   \checkmark       &                &\checkmark       &ResNet-50          &42.89     \\
   \checkmark       &\checkmark      &\checkmark       &ResNet-50          &44.09     \\
   \hline
   \hline
\end{tabular}}
\end{table}

\begin{table}[t]
\centering
\caption{
   Complexity comparison with existing context schemes. 
   The feature map of size $[1 \times 2048 \times 128 \times 128]$ is adopted to evaluate their complexity during inference.
   All the numbers are obtained on a single NVIDIA Tesla V100 GPU with CUDA 11.0 and the smaller, the better.
   As seen, our method requires the least Parameters and the least FLOPs.
}\label{table2}
\resizebox{.45\textwidth}{!}{
\begin{tabular}{c|c|c|c}
   \hline
   \hline
   Method                                              &Parameters      &FLOPs            &Time                  \\
   \hline
   ASPP \cite{chen2017rethinking} (\emph{our impl.})   &42.21M          &674.47G          &101.44ms              \\
   PPM \cite{zhao2017pyramid} (\emph{our impl.})       &23.07M          &309.45G          &29.57ms               \\
   CCNet \cite{huang2019ccnet} (\emph{our impl.})      &23.92M          &397.38G          &56.90ms               \\
   OCRNet \cite{yuan2019object} (\emph{our impl.})     &14.82M          &237.45G          &20.22ms               \\
   DANet \cite{fu2019dual} (\emph{our impl.})          &23.92M          &392.02G          &62.64ms               \\
   ANN \cite{zhu2019asymmetric} (\emph{our impl.})     &20.32M          &335.24G          &49.66ms               \\
   DNL \cite{yin2020disentangled} (\emph{our impl.})   &24.12M          &395.25G          &68.62ms               \\
   APCNet \cite{he2019adaptive} (\emph{our impl.})     &30.46M          &413.12G          &54.20ms               \\
   \hline
   ILCM (\emph{ours})                                  &10.36M          &169.77G          &42.56ms               \\
   SLCM (\emph{ours})                                  &10.10M          &165.47G          &53.12ms               \\
   ILCM+SLCM (\emph{ours})                             &11.02M          &180.60G          &84.19ms                \\
   \hline
   \hline
\end{tabular}}
\vspace{-0.30cm}
\end{table}

\noindent \textbf{SLCM.} 
Since the label of a pixel is essentially the category of the object the pixel belongs to, 
the contextual information from the same semantic class could further enhance the category representation ability of the original pixel representations.
Accordingly, the network can leverage the enhanced representations to classify the pixels more accurately.
As demonstrated in Table \ref{table1}, we can see that aggregating the semantic-level contextual information can improve the performance of base framework by $5.93\%$ in terms of  mIoU.
This improvement well demonstrates the effectiveness of the proposed semantic-level context module.

\noindent \textbf{ILCM+SLCM.} 
Co-occurrent visual pattern makes image-level contextual information be critical for semantic segmentation.
However, capturing the contextual information from the whole image will also cause some problems.
For instance, since the label of a pixel is decided by the category of the object the pixel belongs to, 
aggregating too much contextual information from other category regions may cause the network to mislabel the pixel as other categories.
To address this issue, this paper proposes to introduce the semantic-level contextual information for each pixel representation additionally,
which only leverages the representations in the corresponding category region to enhance each pixel representation.
As illustrated in Table \ref{table1}, we can find that combining ILCM and SLCM outperforms the base model by $7.13\%$ in terms of mIoU.
The improvement is much higher than applying single ILCM ($7.13\%$ \emph{v.s.} $5.54\%$) or single SLCM ($7.13\%$ \emph{v.s.} $5.93\%$).
This result indicates that ILCM and SLCM can complement and promote each other, 
which well demonstrates the reliability of the basic motivation, and the effectiveness of the designed framework in this paper.

\begin{table}[t]
\centering
\caption{
   Comparison with the existing context schemes in terms of mIoU. 
   All the models here are learned on the train set of ADE20K and tested on the validation set of ADE20K.
}\label{table3}
\resizebox{.45\textwidth}{!}{
\begin{tabular}{c|c|c|c|c}
   \hline
   \hline
   Method                                            &Backbone   &Stride          &Iterations  &mIoU              \\
   \hline
   PPM \cite{zhao2017pyramid} (\emph{our impl.})     &ResNet-50  &$8\times$       &160K        &42.64               \\
   ASPP \cite{chen2017rethinking} (\emph{our impl.}) &ResNet-50  &$8\times$       &160K        &43.19               \\
   OCR \cite{yuan2019object} (\emph{our impl.})      &ResNet-50  &$8\times$       &160K        &42.47               \\
   ANN \cite{zhu2019asymmetric} (\emph{our impl.})   &ResNet-50  &$8\times$       &160K        &41.75               \\
   NonLocal \cite{wang2018non} (\emph{our impl.})    &ResNet-50  &$8\times$       &160K        &42.15               \\
   DNL \cite{yin2020disentangled} (\emph{our impl.}) &ResNet-50  &$8\times$       &160K        &43.50               \\
   CCNet \cite{huang2019ccnet} (\emph{our impl.})    &ResNet-50  &$8\times$       &160K        &42.47               \\
   \hline
   ISNet (\emph{ours})                               &ResNet-50  &$8\times$       &160K        &\textbf{44.09}       \\
   \hline
   \hline
\end{tabular}}
\end{table}

\begin{figure}
\centering
\includegraphics[width=0.45\textwidth]{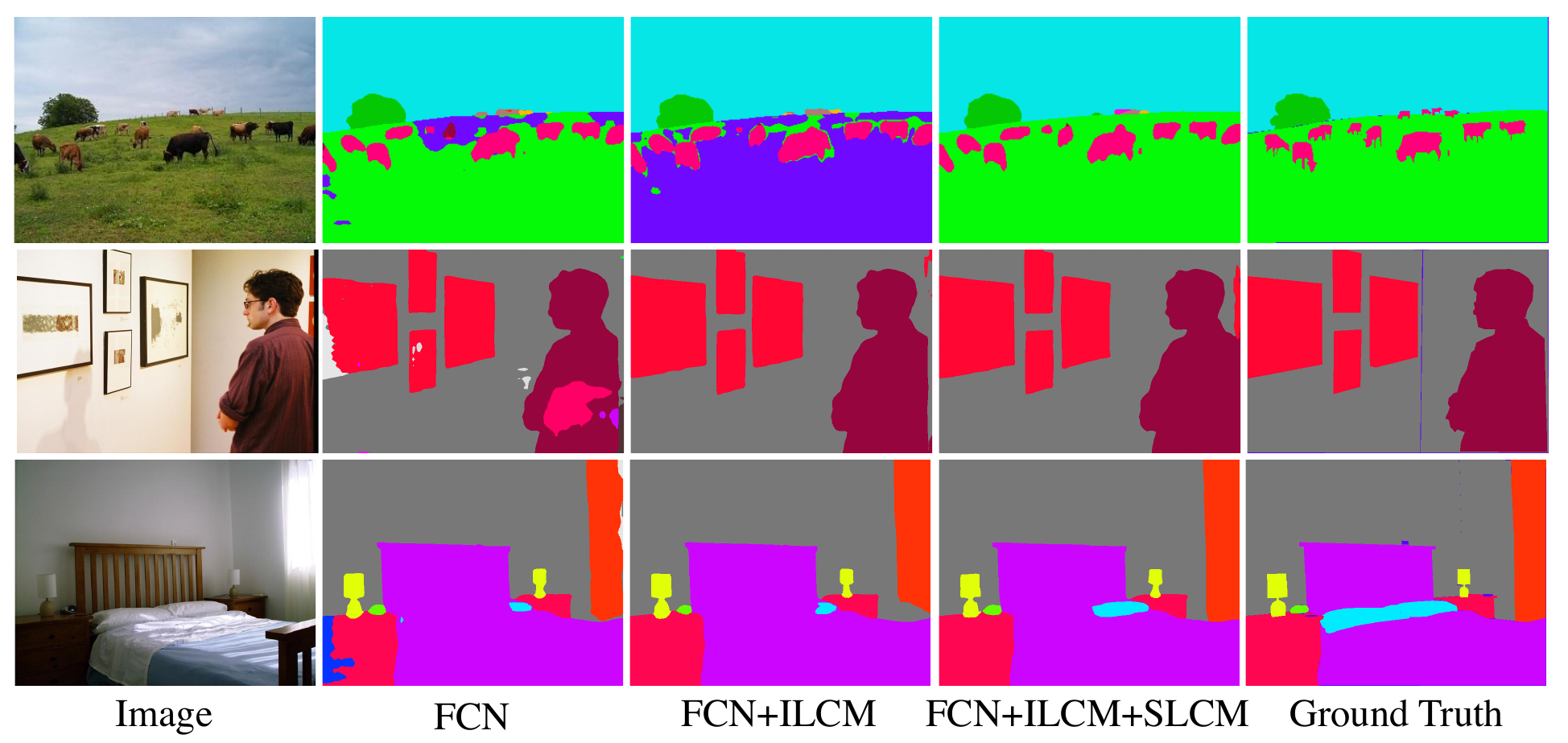}
\caption{
   Qualitative results on the validation set of ADE20K. 
   All the models here are trained under the same setting with ResNet-50 as the backbone network. 
   Best viewed in color and zoom in.
}\label{qr}
\vspace{-0.30cm}
\end{figure}

\noindent \textbf{Qualitative Results.} 
Figure \ref{qr} shows some qualitative results to further prove the reliability of our basic motivation.
As seen, the segmentation performance has been well improved after introducing the semantic-level context module (SLCM).
For instance, in the 2nd row, the painting with only a small visible part on the right side of the image is still well segmented (the ground truth is wrong).
This result well indicates that aggregating too much contextual information from other category regions may cause the network to mislabel one pixel as other categories, 
and introducing SLCM can well alleviate this problem.

\noindent \textbf{Complexity.} 
Table \ref{table2} demonstrates the complexity comparison with existing context schemes, including the increased parameters, computation complexity (measured by the number of FLOPs) and inference time.
As seen, the proposed context scheme requires the least parameters and the least computation complexity.
Specifically, ILCM+SLCM only requires $\frac{1}{4}$ and $\frac{1}{2}$ of the parameters of ASPP and PPM respectively, which can prevent our model from overfitting to a certain extent.
Furthermore, ILCM+SLCM only requires $\frac{1}{2}$, $\frac{1}{4}$, $\frac{1}{2}$, $\frac{7}{10}$, $\frac{1}{2}$, $\frac{1}{2}$, $\frac{1}{2}$, $\frac{2}{5}$ of the FLOPs based on PPM, ASPP, CCNet, OCRNet, DANet, ANNet, DNL and APCNet respectively.
These results well prove the efficiency of the proposed method.

\noindent \textbf{Performance Comparison.} 
To further show the necessity of introducing the semantic-level contextual information, we compare the performance of ISNet with the existing context schemes under the same training and testing settings.
As illustrated in Table \ref{table3}, we can see that ISNet outperforms all the existing context modules by yielding a mIoU of $44.09\%$.
Noted that, the image-level context module integrated into ISNet only gains a mIoU of $42.50\%$ which is much weaker than most of the existing image-level context schemes.
This result well demonstrates the effectiveness of introducing the semantic-level contextual information.

\begin{table}[t]
\centering
\caption{
   Segmentation results on ADE20K validation set. 
   Multi-scale and flipping testing is employed here for fair comparison.
   The best score is marked in \textbf{bold}.
}\label{table4}
\resizebox{.4\textwidth}{!}{
\begin{tabular}{c|c|c|c}
   \hline
   \hline
   Method                                           &Backbone      &Stride       &mIoU              \\
   \hline
   PSPNet \cite{zhao2017pyramid}                    &ResNet-101    &$8\times$    &43.29             \\
   PSANet \cite{zhao2018psanet}                     &ResNet-101    &$8\times$    &43.77             \\
   EncNet \cite{zhang2018context}                   &ResNet-101    &$8\times$    &44.65             \\
   OCNet \cite{yuan2018ocnet}                       &ResNet-101    &$8\times$    &45.08             \\
   OCRNet \cite{yuan2019object}                     &ResNet-101    &$8\times$    &45.28             \\
   CCNet \cite{huang2019ccnet}                      &ResNet-101    &$8\times$    &45.76             \\
   ANNet \cite{zhu2019asymmetric}                   &ResNet-101    &$8\times$    &45.24             \\
   ACNet \cite{fu2019adaptive}                      &ResNet-101    &$8\times$    &45.90             \\
   DMNet \cite{he2019dynamic}                       &ResNet-101    &$8\times$    &45.50             \\
   APCNet \cite{he2019adaptive}                     &ResNet-101    &$8\times$    &45.38             \\
   DANet \cite{fu2019dual}                          &ResNet-101    &$8\times$    &45.22             \\
   OCRNet \cite{yuan2019object}                     &HRNetV2-W48   &$4\times$    &45.66             \\
   \hline
   ISNet (\emph{ours})                              &ResNet-50     &$8\times$    &45.04            \\
   ISNet (\emph{ours})                              &ResNet-101    &$8\times$    &47.31            \\
   ISNet (\emph{ours})                              &ResNeSt-101   &$8\times$    &\textbf{47.55}   \\
   \hline
   \hline
\end{tabular}}
\vspace{-0.3cm}
\end{table}

\begin{table}[t]
\centering
\caption{
   Comparison of performance on the test set of COCOStuff with state-of-the-art approaches.
   Multi-scale and flipping testing is leveraged here for fair comparison.
}\label{table5}
\resizebox{.4\textwidth}{!}{
\begin{tabular}{c|c|c|c}
   \hline
   \hline
   Method                                           &Backbone      &Stride       &mIoU              \\
   \hline
   OCRNet \cite{yuan2019object}                     &ResNet-101    &$8\times$    &39.50             \\
   SVCNet \cite{ding2019semantic}                   &ResNet-101    &$8\times$    &39.60             \\
   DANet \cite{fu2019dual}                          &ResNet-101    &$8\times$    &39.70             \\
   EMANet \cite{li2019expectation}                  &ResNet-101    &$8\times$    &39.90             \\
   SpyGR \cite{li2020spatial}                       &ResNet-101    &$8\times$    &39.90             \\
   ACNet \cite{fu2019adaptive}                      &ResNet-101    &$8\times$    &40.10             \\
   OCRNet \cite{yuan2019object}                     &HRNetV2-W48   &$4\times$    &40.50             \\
   \hline
   ISNet (\emph{ours})                              &ResNet-50     &$8\times$    &40.16             \\
   ISNet (\emph{ours})                              &ResNet-101    &$8\times$    &41.60             \\
   ISNet (\emph{ours})                              &ResNeSt-101   &$8\times$    &\textbf{42.08}    \\
   \hline
   \hline
\end{tabular}}
\vspace{-0.3cm}
\end{table}

\subsection{Comparison with State-of-the-Art}
\noindent \textbf{ADE20K.} 
Results of other state-of-the-art semantic segmentation solutions on ADE20K are summarized in Table \ref{table4}.
As is known, ADE20K is challenging due to its various image scales, plenty of semantic classes and the gap between its training and validation set.
Even under such circumstance, ISNet employing ResNet-50 achieves a mIoU of $45.04\%$, 
which is $1.75\%$, $1.27\%$ and $0.39\%$ mIoU higher than PSPNet \cite{zhao2017pyramid}, PSANet \cite{zhao2018psanet} and EncNet \cite{zhang2018context} using a stronger ResNet-101 backbone network, respectively.
This result further shows the importance of aggregating the semantic-level contextual information for augmenting each pixel representation.
Furthermore, as we can see, the previous best method named ACNet achieves a mIoU of $45.90\%$. 
Our ISNet with ResNet-101 achieves superior mIoU of $47.31\%$ which is $1.41\%$ mIoU higher than the previous state-of-the-art.
Besides, integrating ILCM and SLCM also allows this paper to report new state-of-the-art performance on the validation set of ADE20K, \emph{i.e.}, $47.55\%$ by leveraging ResNeSt-101.

\begin{table}[t]
\centering
\caption{
   State-of-the-art comparison on the validation set of LIP.
   Flipping testing is utilized here for fair comparison.
   $\ddagger$ means that we adopt ASPP as the image-level context module.
}\label{table6}
\resizebox{.4\textwidth}{!}{
\begin{tabular}{c|c|c|c}
   \hline
   \hline
   Method                                           &Backbone      &Stride       &mIoU              \\
   \hline
   DeepLab \cite{chen2017deeplab}                   &ResNet-101    &-            &44.80             \\
   CE2P \cite{ruan2019devil}                        &ResNet-101    &$16\times$   &53.10             \\
   OCRNet \cite{yuan2019object}                     &ResNet-101    &$8\times$    &55.60             \\
   OCNet \cite{yuan2018ocnet}                       &ResNet-101    &$8\times$    &54.72             \\
   CCNet \cite{huang2019ccnet}                      &ResNet-101    &$8\times$    &55.47             \\
   HRNet \cite{wang2020deep}                        &HRNetV2-W48   &$4\times$    &55.90             \\
   OCRNet \cite{yuan2019object}                     &HRNetV2-W48   &$4\times$    &56.65             \\
   \hline
   ISNet (\emph{ours})                              &ResNet-50     &$8\times$    &53.41             \\
   ISNet (\emph{ours})                              &ResNet-101    &$8\times$    &55.41             \\
   ISNet (\emph{ours})                              &ResNeSt-101   &$8\times$    &56.81             \\
   ASPP (\emph{our impl.})                          &ResNet-101    &$8\times$    &55.34             \\
   ISNet$\ddagger$ (\emph{ours})                    &ResNet-101    &$8\times$    &\textbf{56.96}    \\
   \hline
   \hline
\end{tabular}}
\vspace{-0.30cm}
\end{table}

\begin{table}[t]
\centering
\caption{
   Segmentation results on Cityscapes validation set. 
   Only single-scale testing is adopted here.
}\label{table7}
\resizebox{.4\textwidth}{!}{
\begin{tabular}{c|c|c|c}
   \hline
   \hline
   Method                                           &Backbone      &Stride               &mIoU       \\
   \hline
   GCNet  \cite{cao2019gcnet,mmseg2020}             &ResNet-101     &$8\times$             &79.03      \\
   PSPNet \cite{zhao2017pyramid,mmseg2020}          &ResNet-101     &$8\times$             &79.76      \\
   PSANet  \cite{zhao2018psanet,mmseg2020}          &ResNet-101     &$8\times$             &79.31      \\
   ANN \cite{zhu2019asymmetric,mmseg2020}           &ResNet-101     &$8\times$             &77.14     \\
   NonLocal  \cite{wang2018non,mmseg2020}           &ResNet-101     &$8\times$             &78.93      \\
   CCNet  \cite{huang2019ccnet,mmseg2020}           &ResNet-101     &$8\times$             &78.87      \\
   EncNet \cite{zhang2018context,mmseg2020}         &ResNet-101     &$8\times$             &78.55      \\
   DANet \cite{fu2019dual,mmseg2020}                &ResNet-101     &$8\times$             &80.41	   \\
   DNL \cite{yin2020disentangled,mmseg2020}         &ResNet-101     &$8\times$             &80.41	   \\	
   OCRNet \cite{yuan2019object,mmseg2020}           &HRNetV2-W48    &$4\times$             &80.70      \\
   \hline
   ISNet (\emph{ours})                              &ResNet-50      &$8\times$             &79.32      \\
   ISNet (\emph{ours})                              &ResNet-101     &$8\times$             &80.56             \\
   ISNet$\ddagger$ (\emph{ours})                    &ResNet-101     &$8\times$             &\textbf{81.10}       \\
   \hline
   \hline
\end{tabular}}
\vspace{-0.30cm}
\end{table}

\begin{figure*}
\centering
\includegraphics[width=0.90\textwidth]{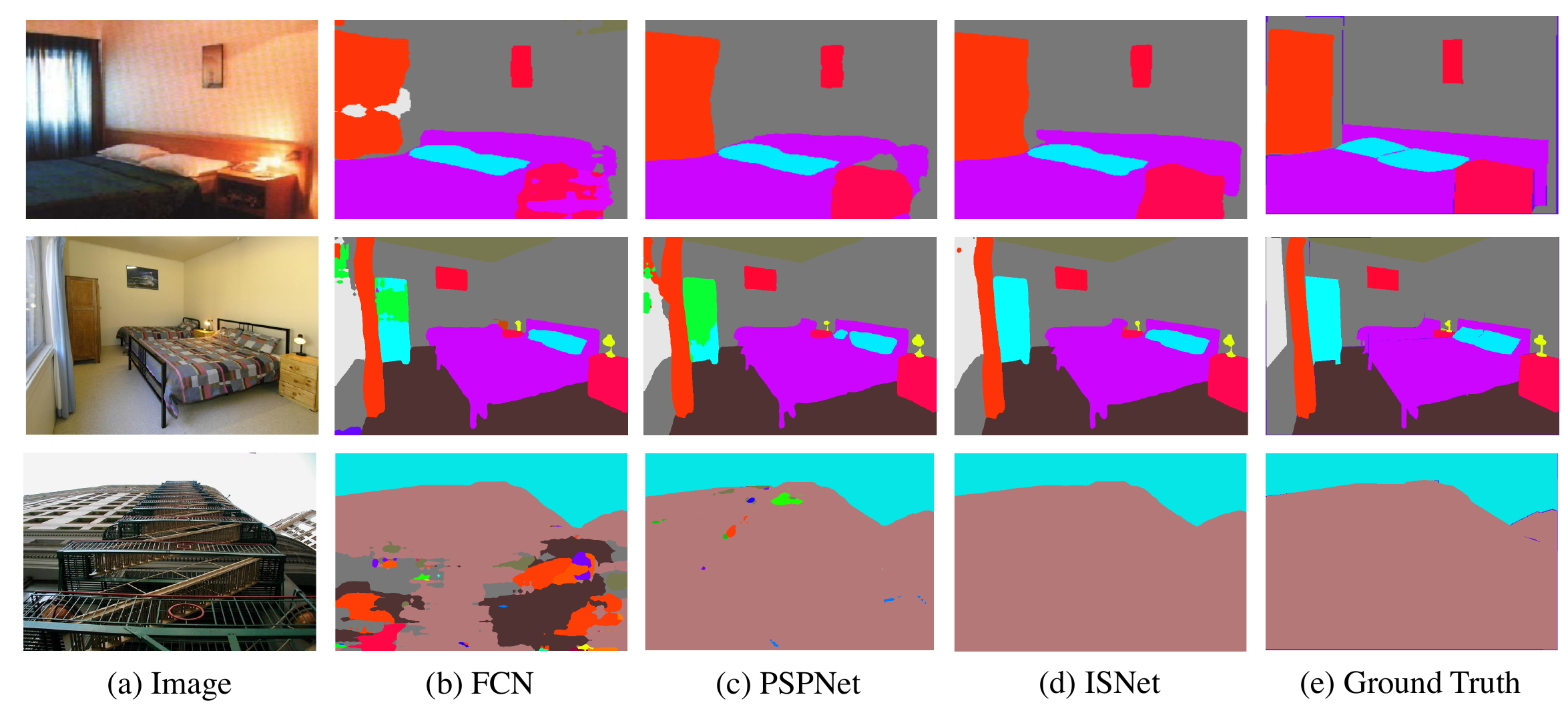}
\caption{
   Qualitative results on the validation set of ADE20K. 
   All the models here are trained under the same setting. 
   Best viewed in color and zoom in.
}\label{compare}
\vspace{-0.50cm}
\end{figure*}

\noindent \textbf{COCOStuff.} 
Since there are only 9K images in the train set of COCOStuff where these images contain 182 semantic classes, COCOStuff is a very challenging benchmark for scene parsing.
Table \ref{table5} compares the performance of the state-of-the-art methods.
By leveraging ResNet-50 as the backbone network, ISNet achieves a mIoU of $40.16\%$, which is already higher than the most of the previous state-of-the-art methods.
When leveraging the same backbone network ResNet-101, our ISNet surpasses OCRNet \cite{yuan2019object}, EMANet \cite{li2019expectation}, DANet \cite{fu2019dual} and ACNet \cite{fu2019adaptive} in a large margin, \emph{i.e.},
$2.10\%$, $1.70\%$, $1.90\%$ and $1.50\%$ mIoU, respectively.
Furthermore, because of the effectiveness of integrating image-level and semantic-level context for semantic image segmentation,
our ISNet with ResNeSt-101 reports the new state-of-the-art performance on the testing set of COCOStuff, \emph{i.e.}, $42.08\%$.

\noindent \textbf{LIP.} 
LIP is a fine-grained semantic segmentation benchmark which has additional challenges such as the complex clothes texture, the scale diversity of different categories, the deformable human body, and the fine-grained segmentation of labels.
Therefore, it is hard to model the image-level context by only leveraging the proposed ILCM, which simply concatenates output features of an average pooling layer. 
Despite this, as demonstrated in Table \ref{table6}, ISNet with ResNet-101 still achieves a mIoU of $55.41\%$ which is very competitive among the previous state-of-the-art methods.
To report a new state-of-the-art performance, we replace the original ILCM with ASPP \cite{chen2017deeplab} here to better model the image-level context.
As seen, our ISNet$\ddagger$ outperforms the previous best method OCRNet with HRNetV2-W48 by $0.31\%$ mIoU while ASPP only achieves $55.34\%$ in terms of mIoU which is $1.62\%$ lower than our ISNet$\ddagger$.
These results consistently prove the basic motivation of this paper, \emph{i.e.}, additionally introducing the semantic-level contextual information is critical to improving the pixel representations.

\noindent \textbf{Cityscapes.} 
As shown in Table \ref{table7},
we also show the comparative results with other state-of-the-art methods on the validation set of Cityscapes.
We can see that our model ISNet$\ddagger$ with ResNet-101 is superior to previous best method OCRNet with HRNetV2-W48.
Specifically, integrating the image-level context module ASPP and the proposed semantic-level context module SLCM makes this paper report a new state-of-the-art with mIoU hitting $81.10\%$ under single-scale testing. 
This result further proves the rationality of aggregating the image-level and semantic-level context for augmenting each pixel representation.

\noindent \textbf{Qualitative Results.} 
Figure \ref{compare} illustrates the qualitative results on the validation set of ADE20K.
We can see that our ISNet could achieve better segmentation results than both FCN (\emph{i.e.}, without context module) and PSPNet (\emph{i.e.}, using the image-level context module), 
which further shows the effectiveness of our method (\emph{i.e.}, adopting both image-level and semantic-level context module).

\section{Conclusion}
This paper studies the context aggregation problem.
Motivated by the fact that the existing image-level context schemes may bring too much contextual information of other categories into the pixel representations so that it makes the network mislabel the pixel,
we propose to integrate the image-level contextual information and semantic-level contextual information, respectively, to further boost the performance of semantic segmentation.
Specifically, we first design a simple yet effective image-level context module as a common practice to capture the global semantic structured information.
Then, the semantic-level contextual information is also aggregated for each pixel by leveraging the proposed semantic-level context module.
At last, the pixel representations are augmented by weighted aggregating the image-level contextual information and the semantic-level contextual information.
Extensive experiments demonstrate the effectiveness of our method.
Integrating image-level and semantic-level context allows us to report new-state-of-arts on four segmentation benchmarks, \emph{i.e.}, ADE20K, Cityscapes, LIP and COCOStuff.

{\small
\bibliographystyle{ieee_fullname}
\bibliography{egbib}
}

\end{document}